# Human Leg Motion Tracking by Fusing IMUs and RGB Camera Data Using Extended Kalman Filter

Omid Taheri [1], Hassan Salarieh [2] and Aria Alasty [3]

[1] Affiliation 1; omid.taheri@tuebingen.mpg.de
[2] Affiliation 2; salarieh@sharif.edu
[3] Affiliation 3; aalasti@sharif.edu

**Abstract:** Human motion capture is frequently used to study rehabilitation and clinical problems, as well as to provide realistic animation for the entertainment industry. IMU-based systems, as well as Marker-based motion tracking systems, are most popular methods to track movement due to their low cost of implementation and lightweight. This paper proposes a quaternion-based Extended Kalman filter approach to recover the human leg segments motions with a set of IMU sensors data fused with camera-marker system data. In this paper, an Extended Kalman Filter approach is developed to fuse the data of two IMUs and one RGB camera for human leg motion tracking. Based on the complementary properties of the inertial sensors and camera-marker system, in the introduced new measurement model, the orientation data of the upper leg and the lower leg is updated through three measurement equations. The positioning of the human body is made possible by the tracked position of the pelvis joint by the camera marker system. A mathematical model has been utilized to estimate joints' depth in 2D images. The efficiency of the proposed algorithm is evaluated by an optical motion tracker system.

**Keywords:** Motion Capture, EKF, Data fusion, Marker-based systems, IMU sensor, Rehabilitation

## *1. Introduction*

Human motion tracking (Motion capture) plays a critical role in a wide range of fields including Health-care services[1], Virtual and Augmented reality, entertainments, and sports. In Health-care services, the result of motion tracking is applied for diagnosis and treatment of patients. In entertainment, motion capture is used to animate a 3D character in movies and games while sports use it for injury prevention and improving performances [2]. Different applications and situations had led to the development of various motion tracking systems.

Among all the motion capture techniques, camera and marker-based systems are the most prevailing technique which have high accuracy. By attaching a marker to a joint, regardless of the complexity of human motion, a joint location is tracked by tracking the marker location using image processing techniques. However, due to certain limitations like the need for multiple high-resolution cameras, restriction to the studio-like environment, a huge amount of data, and high cost, the motivation to use these systems have been diminished. Nowadays, due to the recent advances in MEMS sensors, low cost, small size, light weight, and low energy consumption of inertial sensors make them more attractive for the research on human motion capture. However, these sensors have their own restrictions.

Inertial sensors (IMUs) consist of three orthogonal gyroscopes, accelerometers, and magnetometers. In IMU-based systems, a sensor is attached to one body segment. By fusing sensory data, segment orientation can be estimated. Based on the estimated orientation, together with the length of each segment and the arranging relationship between segments, the motion of the whole body can be obtained. These systems have no line-of-sight requirements, and no emitters to install [3]. Thus, IMU-based systems can be applied in a variety of applications where a studio-like environment is not necessary. In inertial systems, due to the integration of gyroscope signal over time, results lead to drift errors. Another drawback is that IMUs are not well-suited [4] for determining absolute location. For accurate and drift free orientation estimation many fusion algorithms have been reported combining the signals from gyroscopes, accelerometers and magnetometers [5, 6]. The Kalman filter [7] and optimization function have become the accepted basis for the majority of orientation estimation algorithms[8-12].

Regarding the limitations of Camera-based and IMU-based motion capture systems, a fusion of these sensors seems to enhance tracking results. An Extended Kalman Filter approach is proposed by Araguás et al. for visual estimation of a Hovering UAV orientation using IMUs and a camera[13]. IMUs provide information for roll and pitch angles, whereas the camera is used for measuring yaw angle. Bo et al. explored the fusion of IMUs and Kinect [4] in rehabilitation for 1DOF joint motions, e.g., knee flexion and extension. Using Kalman Filter, the angle estimation from the inertial sensors is corrected whenever Kinect data is present. An unscented Kalman filter method is used by Tian et al. [14] to fuse IMU and Kinect data for human upper limb motion tracking. The position of the elbow and the wrist joint are computed by integrating the accelerometers data twice, then as a geometrical constraint, these positions are fused with Kinect data to compensate for IMUs drift. Destelle et al. developed an algorithm [15] in which the position of the torso joint is obtained from Kinect and then the position of the other joints and the orientation of the segments are calculated by the inertial sensors alone. Feng and Murray-Smith [16] developed a multi-rate Kalman filter method to track the hand position with the fusion of an inertial sensor and Kinect. Atrsaei et al. proposed a fusion algorithm between the Kinect and inertial sensors using an unscented Kalman filter to track human arm movements[17].

This paper describes the design, implementation, and experimental testing of a human leg motion tracking system. This approach is based on the fusion of leg segments' orientation data, obtained by both IMUs and an RGB camera, using Extended Kalman Filter (EKF). Our goal is to explore the complementary properties and fuse these sensors to improve the system state estimation. We have used a marker-based motion tracking system to track joint locations in the sagittal plate. The depth of joints is also estimated using a mathematical model. In addition, in order to reduce drift of IMUs, the effect of geometrical and kinematical constraints had been explored. The main contributions of this paper are:

1. Using quaternion to represent the rotation and orientation not to suffer from the singularity problem
2. Tracking leg motion during walking, i.e., absolute location of body is calculated by camera and the relative location of joints is calculated by both systems
3. Reducing error by implementing kinematical constraints
4. Using only one camera to track 3D joint location
5. Low cost implementation due to using a simple camera with IMUs

The complete method was then evaluated using a high accuracy optical motion tracking system.

This paper is organized as follows: Section 2 describes the system setup. Section 3 presents leg tracking using a single camera with markers. The structure of the proposed algorithm for tracking by IMUs is described in section 4. Section 5 details the design of Extended Kalman Filter. The process model and measurement equations are also described. Section 6 describes the implementation of the fusion algorithm and analyzing experiment results. Finally, some conclusions are given in section 7.

## 2. System Setup

The pipeline of our proposed motion tracking system consists of three steps. First, joint locations are estimated using an RGB camera by tracking color markers attached to the joints. Second, the IMU data is used to determine the lower body segments' orientation. Finally, the 3D location of joints is estimated by fusing the data of IMUs and camera using EKF and the kinematic constraints of the body segments. Finally, the results are evaluated by a reference high accuracy Vicon Mocap system. Considering the proposed method, two IMUs and three colored markers are attached to leg segments and leg joints (Pelvis, Knee, and Ankle Joints) respectively as shown in Figure 1. Before the analysis it is necessary to define four coordinate systems: 1) north-east-down (NED) coordinate system which is taken as the reference frame, 2) body coordinate system of the inertial sensors which is assumed coinciding with segment coordinate system, 3) camera coordinate system, and 4) the reference system coordinate system. To represent the rotation of coordinate systems, the quaternions has been employed. Detailed formulation and information of quaternions are presented in [18].

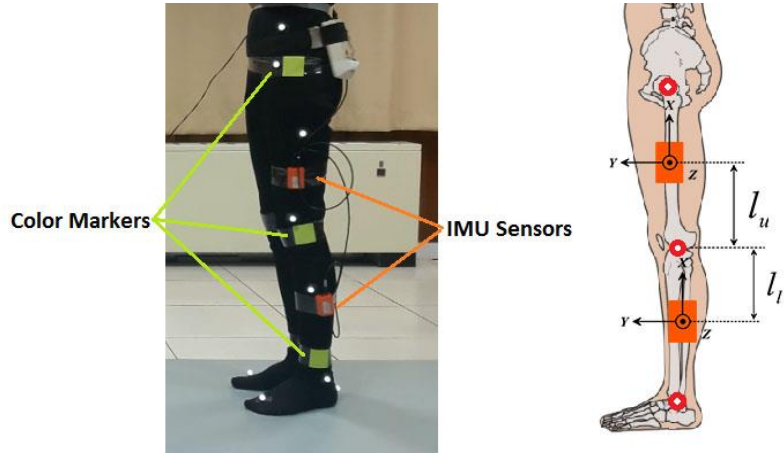

Figure 1. IMU sensors and Markers arrangement on the human leg

## 3. Motion tracking by Camera

### 3.1. 2D Joint Location

Human gait is generally performed in the sagittal plane of the body. In order to achieve better results, considering the use of a single camera, the viewing point of the camera was set perpendicular to the sagittal plane (Figure 2). Joint markers can be of any color and it is assumed that each marker's center is aligned with its joint center and also markers do not move relative to the human skin.

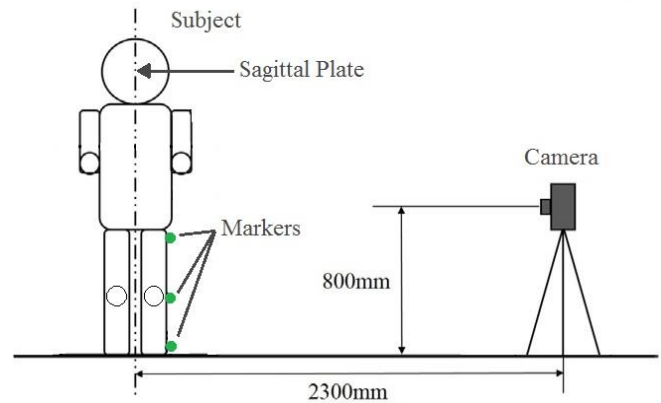

Figure 2. Camera and markers position relative to human body

With an objective of tracking color markers, the sequence of frames from the subject are taken along the optical axis. First, a simple color-based segmentation was used to distinguish color markers in three steps:
1. Convert input RGB frame to grayscale
2. Subtracting gray frame from input RGB frame to remove all colors except marker color
3. Converting subtracted mask to binary mask (only white and black colors)

To recognize markers contour, Connected Contour Labeling (CCL) algorithm [19-21] was used. Next, the 2D location of center and area of each marker contour in the sagittal plane was calculated. All steps are illustrated in Figure 3.

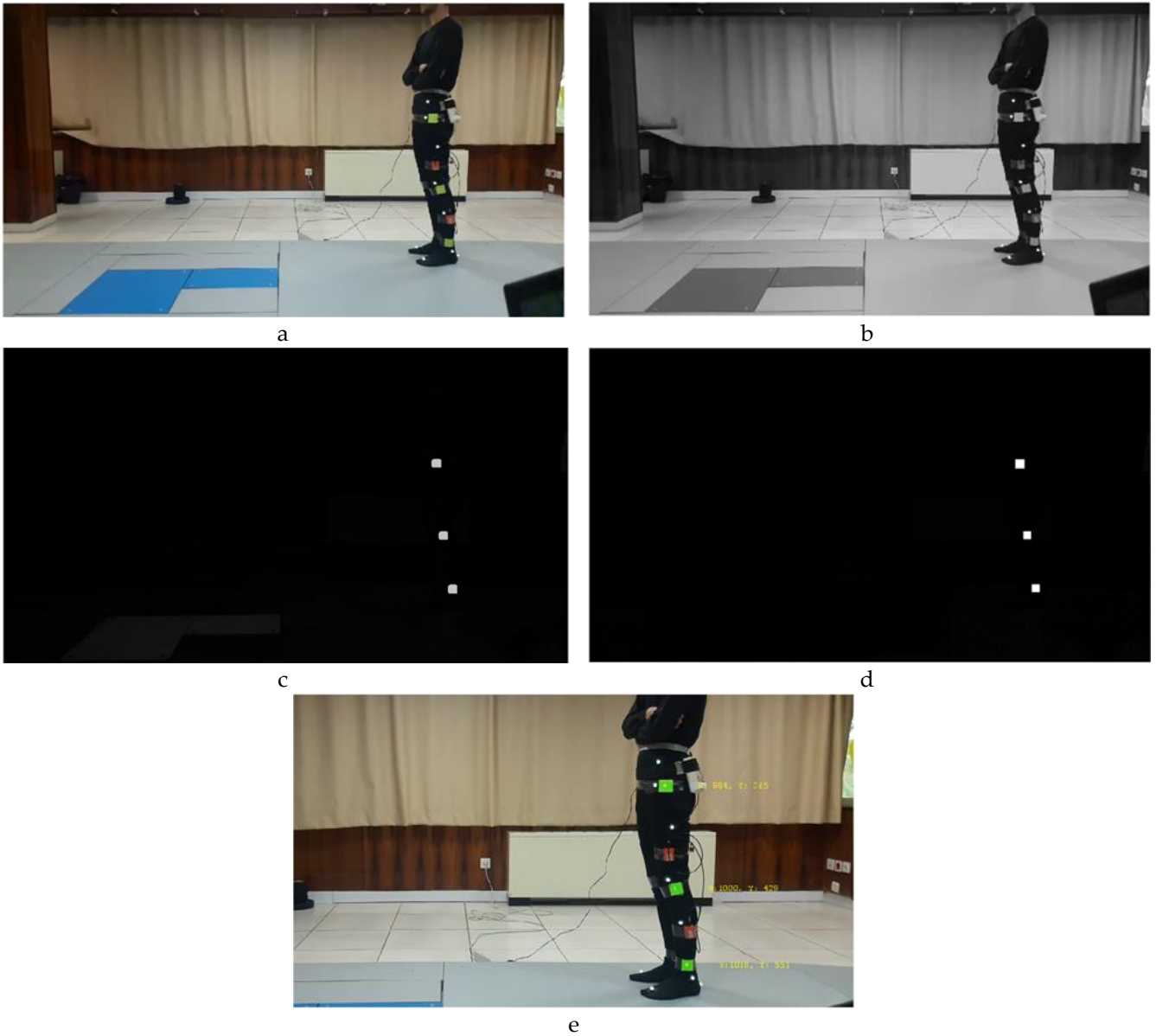

**Figure 3. Joints position tracking steps, a)Input frame, b) gray scale frame, c) marker-color mask, d) binary mask, e)markers 2D position**

### 3.2. Joint Depth Estimation

Depth estimation in 2D images can be accomplished by various algorithms. In this paper, we used a mathematical model proposed by [22] to estimate the depth of each joint. In this method, it is assumed that the optical axis of the camera is perpendicular to object axis. The estimation is based on 1) real area of the object, 2) camera resolution, 3) object pixels in the image, and 4) focal length of the camera.

In this paper, knowing the real distance between markers and the real area of each marker, depth of joints is estimated using these two parameters and then fused by a simple Kalman Filter to enhance the estimation. Due to the accuracy of calculating the distances of markers in relation to their area calculation, the distance-based and area-based equations were chosen as the system model and the measurement equation in Kalman Filter design. Detailed design of Kalman Filter is presented in [10].

## *4. Motion tracking by IMU*

In the proposed algorithm, each body segment's orientation is presented by a quaternion. In order to reduce the drift, the quaternion integrated from gyroscope signals is corrected by the accelerometer and magnetometer signals.

## 4.1. Orientation from Gyroscope

Rate Gyroscopes measure the angular velocity about the x, y and z-axes of the sensor frame, where its quaternion representation $^S\omega$ is defined by Equation (1). The quaternion derivative describing rate of change of the earth frame relative to the sensor frame $\frac{d}{dt}{}^S_N q$ can be calculated [18] by Equation (2) where the operator $\otimes$ denotes a quaternion product. From here by, $^S_N q$ is represented by $q$.

$$^S\omega = [0 \quad \omega_x \quad \omega_y \quad \omega_z], \qquad (1)$$

$$\frac{d}{dt}{}^S_N q = \frac{1}{2}{}^S_N q \otimes {}^S\omega, \qquad (2)$$

To estimate the orientation more accurately by reducing gyroscope bias and noise, it is assumed that the bias is generated by a first order differential equation [15, 23].

$$\frac{d}{dt} b = -Tb + TW, \qquad (3)$$

$$T = \begin{bmatrix} \frac{1}{\tau_1} & 0 & 0 \\ 0 & \frac{1}{\tau_2} & 0 \\ 0 & 0 & \frac{1}{\tau_3} \end{bmatrix}, \qquad (4)$$

where $b$ is the bias vector, $T$ is the time-constant matrix, and $W$ is assumed to be a white Gaussian noise.

Taking component of angular velocity, bias vector, and quaternions as the state vector (Eq. (5)), using Equations (3) and (4) the state-space representation of the system is defined by Equation ((6)).

$$x = \begin{Bmatrix} \omega^u \\ b^u \\ q^u \\ \omega^l \\ b^l \\ q^l \end{Bmatrix}, \qquad (5)$$

$$\dot{x} = f(x, u, w, t), \quad w(t) \sim (0, Q), \qquad (6)$$

where $u$ and $l$ superscripts stand for upper leg and lower leg respectively.

## 4.2. Orientation Correction using Accelerometer and Magnetometer

If the direction of an earth's field is known in the earth frame, a measurement of the field's direction within the sensor frame will allow to calculate the orientation of the sensor frame relative to the earth frame. However, for any given measurement, there will not be a unique sensor orientation solution, instead, there will be infinite solutions represented by all those orientations. This is achieved by the rotation of the true orientation around an axis parallel to the field [18]. The problem of optimal attitude determination algorithm using two vectors (or more), which are known in a reference frame and measured in a moving frame is called in the literature Wahba's problem [24] which has various solution algorithms including Triad, Quest [25], and Optimization [18] algorithm. In the context of an orientation estimation algorithm, it assumes that an accelerometer will measure only gravity and a magnetometer will measure only the earth's magnetic field. Having gravity and magnetic field vectors in both Earth and sensor frames, an optimized orientation of each sensor can be calculated.

In the present case, orientation $_N^S q$ determination is achieved through the formulation of the optimization problem (Eq. ((7)) using Gradient Descent method, proposed by Madgwick et.al in [18].

$$\min_{_N^S\hat{q}\in \Re^4} f(_N^S\hat{q}, {}^N\hat{d}, {}^S\hat{s}), \tag{7}$$

The optimization function is:

$$f(_N^S\hat{q}, {}^N\hat{d}, {}^S\hat{s}) = {}_N^S\hat{q}^* \otimes {}^N\hat{d} \otimes {}_N^S\hat{q} - {}^S\hat{s}, \tag{8}$$

where $_N^S q$ is the orientation quaternion and $^N\hat{d}$ and $^S\hat{s}$ are two vectors in earth and sensor coordinate system respectively. Gradient descent detailed derivation of formulation and equations are given in [18].

Equation (9) calculates the estimated orientation $_N^S q_{\nabla,t}^u$ computed at time $t$ based on a previous estimate of orientation $_N^S q_{\nabla,t-1}^u$ and the objective function error $\nabla f$ defined by sensor measurements of magnetic field ($^S\hat{m}$) and gravity ($^S\hat{a}$) sampled at time t. The subscript $\nabla$ in $_N^S q_{\nabla,t}^u$ indicates that the quaternion is calculated using the gradient descent algorithm.

$$_N^S q_{\nabla,t} = {}_N^S\hat{q}_{est,t-1} - \mu_t \frac{\nabla f}{\|\nabla f\|}, \tag{9}$$

$$\nabla f = \begin{Bmatrix} J_g^T(_N^S\hat{q}_{est,t-1}) f_g(_N^S\hat{q}_{est,t-1}, {}^S a_t) \\ J_{g,b}^T(_N^S\hat{q}_{est,t-1}, {}^N\hat{b}) f_{b,g}(_N^S\hat{q}_{est,t-1}, {}^S\hat{a}, {}^N\hat{b}, {}^S\hat{m}) \end{Bmatrix}, \tag{10}$$

An appropriate value of $\mu_t$ is chosen such that it ensures the convergence rate of $_N^S q_{\nabla,t}^u$ to be limited to the physical orientation rate as this avoids overshooting due to an unnecessarily large step size. Therefore $\mu_t$ can be calculated as in equation (11).

$$\mu_t = \alpha \|_N^S\dot{q}_{\omega,t}\| \Delta t, \alpha > 1, \tag{11}$$

where $\Delta t$ is the sampling period, $_N^S\dot{q}_{\omega,t}$ is the rate of change of orientation measured by gyroscopes and $\alpha$ is an augmentation of $\mu$ to account for noise in accelerometer and magnetometer measurements.

## 5. Kalman Filter Design

It is required to define process model and measurement equations to design Kalman Filter for system data fusion.

### 5.1. Process Model

The process model employed by the filter governs the dynamic relationship between states of two successive time steps [3] so the process model is defined as the state-space equation of the system (Eq. (6)).

### 5.2. Measurement Equation

The measurement model of the system determines the relation between the state variables and the noisy measured parameters from the sensors [17]. To update the measurements of orientation in Kalman Filter, three measurement equations are employed in this paper.

#### 5.2.1. First Measurement Equation

The optimization algorithm is employed to estimate a drift-free orientation of sensors. So, the first measurement equation is:

$$y_{1,k} = h_1(x_k) + v_{k,1} = \begin{Bmatrix} \omega_{m,t}^u \\ {}_N^S q_{\nabla,t}^u \\ \omega_{m,t}^l \\ {}_N^S q_{\nabla,t}^l \end{Bmatrix}, \quad v_{k,1} \sim (0, R_{k,1}), \tag{12}$$

where $\omega_{m,t}$ and ${}_N^S q_{\nabla,t}^u$ are the angular velocity measured by IMUs and gradient decent algorithm orientation estimation measured at the step $k$ respectively.

It is assumed that $v_{k,1}$ is a white gaussian noise with a covariance matrix $R_{k,1}$ [17]:

$$R_{k,1} = \text{diag}[R_{\omega^u} \quad R_{q^u} \quad R_{\omega^l} \quad R_{q^l}], \tag{13}$$

It should be noted that in the gradient descent method, the acceleration of the motion is assumed to be negligible compared to the gravity. Therefore, in high-accelerate motions such as running the covariance matrix components should be set to a large value in order to eliminate first measurement equation. As it is proposed in [4], the threshold of the acceleration can be set to 0.2 in equation ((14).

$$if \; |\|a_{m,t}\| - g| > 0.2 \quad then \quad R_{k,1} = \infty \tag{14}$$

### 5.2.2. Second Measurement Equation

The second equation is a kinematic constraint equation to improve the orientation estimation of the inertial sensors [17]. Equating the knee joint velocity calculated once using lower leg IMU and next through upper leg IMU governs a constraint equation.

Equation (15) represents the absolute velocity of each IMU in NED coordinate system.

$$V_t = V_{t-1} + a_t \, \Delta t = V_{t-1}^l + [q_t \otimes a_{m,t} \otimes q_t^* - \{0 \quad 0 \quad 0 \quad g\}]\Delta t, \tag{15}$$

where $a_{m,t}$ is the acceleration of the sensor. The knee velocity relative to IMU sensors is calculated by:

$$v_k = \omega_k \times \{r \quad 0 \quad 0\}^T, \tag{16}$$

Using Equations ((15) and ((16), the constraint equation governs as follow:

$$V_k^l + [q_k^l \otimes v_k^l \otimes q_k^{l*}] = V_k^u + [q_k^u \otimes v_k^u \otimes q_k^{u*}], \tag{17}$$

Rearranging Equation (17) as a measurement equation, the second measurement equation is obtained.

$$y_{2,k} = h_{2,k}(x_k) + v_{2,k} = 0_{3\times 1}, \quad v_{2,k} \sim (0, R_{2,k}), \tag{18}$$

where $v_{2,k}$ is white Gaussian noise with zero mean. The covariance matrix $R_{2,k}$ is computed using Eq (17).

### 5.2.3. Third Measurement Equation

In this measurement equation, the camera motion tracking data is employed to update the orientation quaternion. Having joint locations, the quaternion of each segment can be calculated using Equation (19).

$$\begin{aligned} {}^C P_{K,t} - {}^C P_{H,t} &= {}_S^C q_t^u \otimes \{0 \quad -l_{u,t} \quad 0 \quad 0\} \otimes {}_S^C q_t^{u*}, \\ {}^C P_{A,t} - {}^C P_{K,t} &= {}_S^C q_t^l \otimes \{0 \quad -l_{l,t} \quad 0 \quad 0\} \otimes {}_S^C q_t^{l*} \end{aligned} \tag{19}$$

where ${}^C P_{K,t}$, ${}^C P_{H,t}$, ${}^C P_{A,t}$ are locations of knee, hip, and ankle joints respectively and ${}_S^C q_t$ calculates the rotation quaternion of the sensor relative to the camera coordinate system.

Thus, the third measurement equation is obtained as:

$$y_{3,k} = h_3(x_k) + v_3 = \begin{Bmatrix} {}^C P_{K,t} - {}^C P_{H,t} \\ {}^C P_{A,t} - {}^C P_{K,t} \end{Bmatrix}, \quad v_{k,3} \sim (0, R_{k,3}), \tag{20}$$

where $v_{3,k}$ is white gaussian noise with zero mean and the covariance matrix $R_{3,k}$ which is computed using Eq. (19).

### 5.3. Filter Design

The process model and measurement equations of the system are both nonlinear, thus an Extended Kalman Filter approach is proposed in this paper. Also, the system is governed by a continuous-time dynamics whereas the measurements are obtained at discrete instants of time, this makes the Hybrid Extended Kalman Filter suitable for the system. Considering the process model and measurement equations of the system, the Hybrid Extended Kalman Filter is designed in seven steps:

1. Initialize filter as follow:

$$\hat{x}_0^+ = \begin{Bmatrix} \boldsymbol{\omega}_{m,1}^u \\ 0_{3\times1} \\ {}_N^S\boldsymbol{q}_{\nabla,0}^u \\ \boldsymbol{\omega}_{m,1}^l \\ 0_{3\times1} \\ {}_N^S\boldsymbol{q}_{\nabla,0}^l \end{Bmatrix},$$

(21)

$$P_0^+ = R_{0,1},$$

2. For $k = 1,2,\ldots$, perform the following steps.
   A. Compute the following partial derivation matrices at the current state estimate:

   $$F = \frac{\partial f}{\partial x}\Big|_{\hat{x}},$$

   $$L = \frac{\partial f}{\partial w}\Big|_{\hat{x}},$$

   (22)

   B. Compute the state estimate and its covariance:

   $$\dot{\hat{x}} = f(\hat{x}, u, 0, t),$$

   $$\dot{P} = FP + PF^T + LQL^T,$$

   (23)

   C. Integrate the state estimation and its covariance to predict the state estimate as below:

   $$\hat{x}_k^- = \int_{t_{k-1}}^{t_k} \dot{\hat{x}} \, dt,$$

   $$P_k^- = \int_{t_{k-1}}^{t_k} \dot{P} \, dt,$$

   (24)

   D. For $i = 1,2,\ldots$, compute the partial derivation matrices using ith measurement equation:

   $$H_{k,i} = \frac{\partial h_{k,i}}{\partial x}\Big|_{\hat{x}_k^-},$$

   $$M_{k,i} = \frac{\partial h_{k,i}}{\partial v}\Big|_{\hat{x}_k^-},$$

   (25)

   E. Finally, compute the Kalman Filter gain and the state estimate correction and its covariance matrix:

   $$K_{k,i} = P_k^- H_k^T (H_k P_k^- H_k^T + M_k R_k M_k^T)^{-1},$$

   $$\hat{x}_{k,i}^+ = \hat{x}_k^- + K_{k,i}[y_{k,i} - h_{k,i}(\hat{x}_k^-, 0)],$$

   $$P_{k,i}^+ = (I - K_{k,i} H_{k,i}) P_{k,i}^-$$

   (26)

## 6. Experimental Results

### 6.1. Setup and Equipment

The algorithm was tested using the Xsens MTx orientation sensor containing 16-bit resolution tri-axis gyroscopes, accelerometers, and magnetometers. Raw sensor data was logged to a PC at 100 Hz and the software provides calibrated sensor measurements which were then processed by the proposed orientation estimation algorithm. For the camera marker system, a Sperado camera with a resolution of 640*480 and a frequency of 30Hz was used by Green square markers with a side length of 5cm which is illustrated in Figure 4.

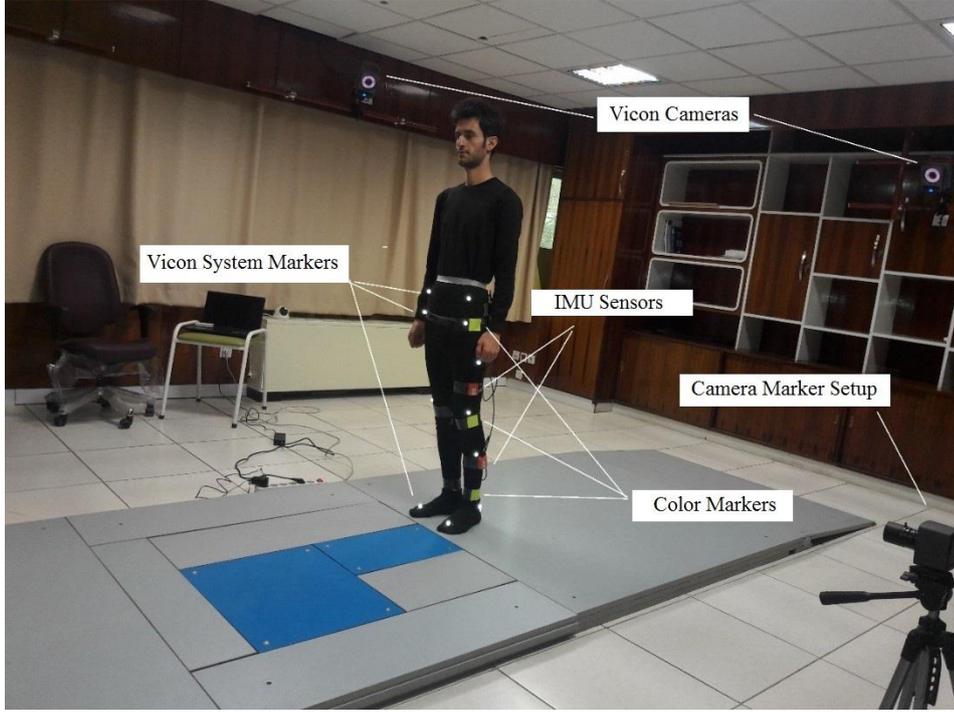

Figure 4. Experiment Setup

In experiments, the Vicon motion tracker system (Vicon Motion Systems, Ltd., Oxford, UK), consisting of 8 MX3+ cameras connected to an MX ultranet server and Nexus software, was used to provide reference measurements of the leg movements tracking. In order to fuse IMU sensor data with the camera-marker system and also for the measurements of the joints position in the Vicon system coordinate frame to be comparable to the estimation algorithm of orientation in the earth frame, it is required to find the rotation quaternion between these coordinate systems. Knowing magnetic field and gravity of earth in both IMU sensor and camera, using a pendulum and magnetic compass, the rotation between both IMUs and camera relative to NED system was obtained using the optimisation algorithm explained earlier. Having the rotation quaternion between NED system and Vicon, IMUs, and camera, the other rotations were computed using Equation (27).

$$^V_S q = ^V_N q \otimes ^N_S q, \qquad (27)$$

The $Q$ and $R_{k,2}$ matrices components were determined by trial and error. Changes in the $Q$ matrix will not have a major effect on the results; however, it could affect the angular velocities by the measurements. By calculating IMUs and camera noise standard deviation during a static posture, the values of the $R_{k,1}$ and $R_{k,3}$ matrices were determined.

$$Q = \mathrm{diag}[0_{1\times 3} \quad 0.045 \quad 0.045 \quad 0.045 \quad 0_{1\times 7} \quad 0.045 \quad 0.045 \quad 0.045 \quad 0_{1\times 4}], \tag{28}$$

$$R_{k,1} = 10^{-3}\mathrm{diag}[0.134 \quad 0.167 \quad 0.035 \quad 0.005 \quad 0.007 \quad 0.002 \quad 0.019 \quad 0.185 \quad 0.029 \quad 0.057 \quad 0.014 \quad 0.004 \quad 0.014 \quad 0.004], \tag{29}$$

$$R_{k,2} = 10^{-4}\mathrm{diag}[1 \quad 1 \quad 1], \tag{30}$$

$$R_{k,3} = \mathrm{diag}[10^{-8} \quad 10^{-8} \quad 10^{-8} \quad 10^{-8} \quad 10^{-8} \quad 10^{-8}], \tag{31}$$

### 6.2. Results and Discussion

For the evaluation of the proposed algorithm, two representative experiments were carried out. In the first one, the subject was performing low-acceleration tasks like walking, which allows the use of the first measurement equation in the fusion algorithm. In the second one, a high-acceleration task like running or jumping was performed by the subject to test the performance of the algorithm in the presence of acceleration.

#### 6.2.1. First Experiment

In the first stage, the subject was asked to stand stationary for few seconds then walk in normal speed along laboratory three times back and forth. As it is illustrated in Figure 5 the acceleration is in a range that the accelerometer and magnetometer signal could be used to correct orientation of sensors.

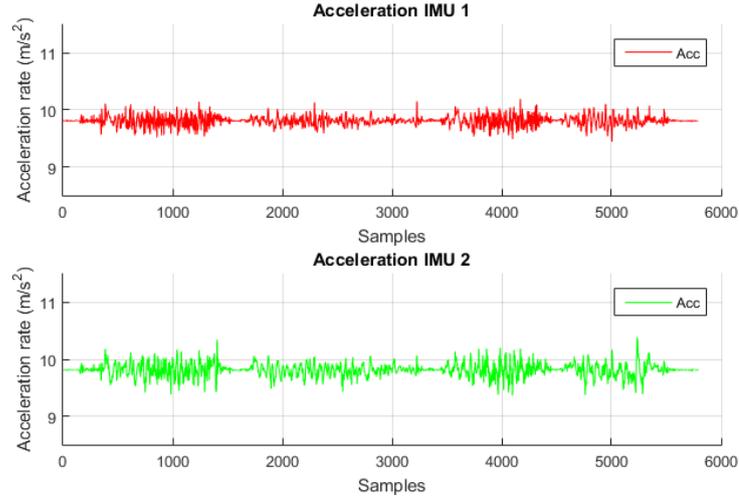

Figure 5. IMUs' acceleration magnitude during first experiment

In Figure 6 and Figure 7 results of IMU sensors and camera-marker motion tracking without fusion are shown respectively. The fusion results for the location of the pelvis, knee, and ankle joints in comparison to the Vicon system are shown in Figure 8. In motion estimation using only IMU sensors, considering that these sensors are not able to measure the absolute location, the absolute location of knee and ankle joint are calculated by IMU sensors data using the absolute location of pelvis calculated by the Vicon system. The accuracy of each algorithm results is summarized in Table 1, Table 2, Table 4, and Table 5, for the first experiment. These tables show the mean RMSEs of joints location. Also, in Table 3 the RMSE of joint depth estimation using the mathematical model is shown.

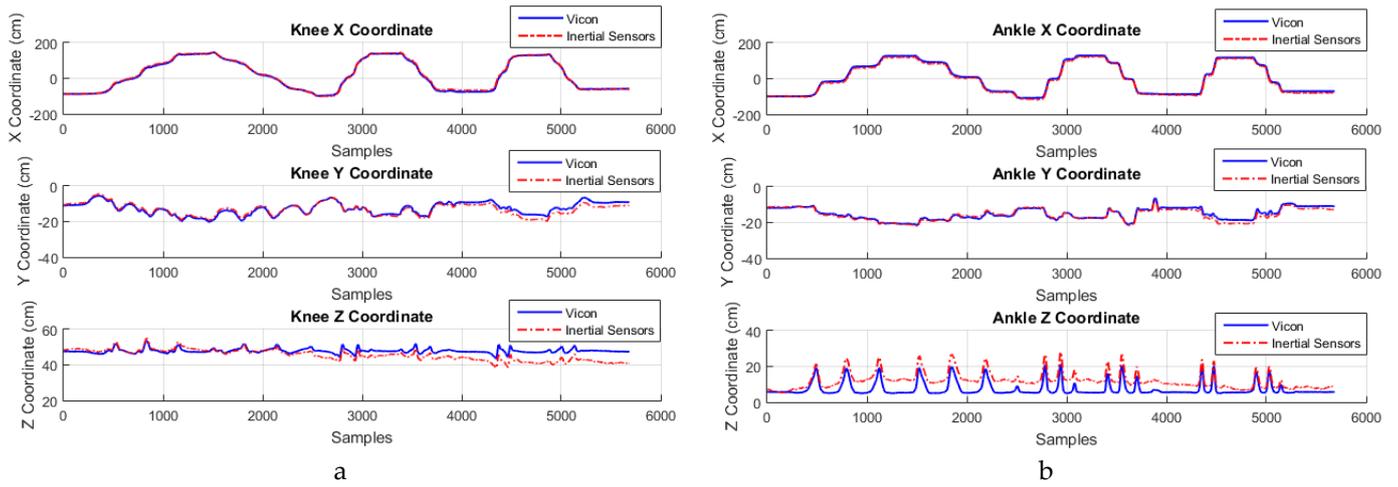

Figure 6. The results of a)Knee, b)Ankle joint motion tracking using IMUs only – First experiment

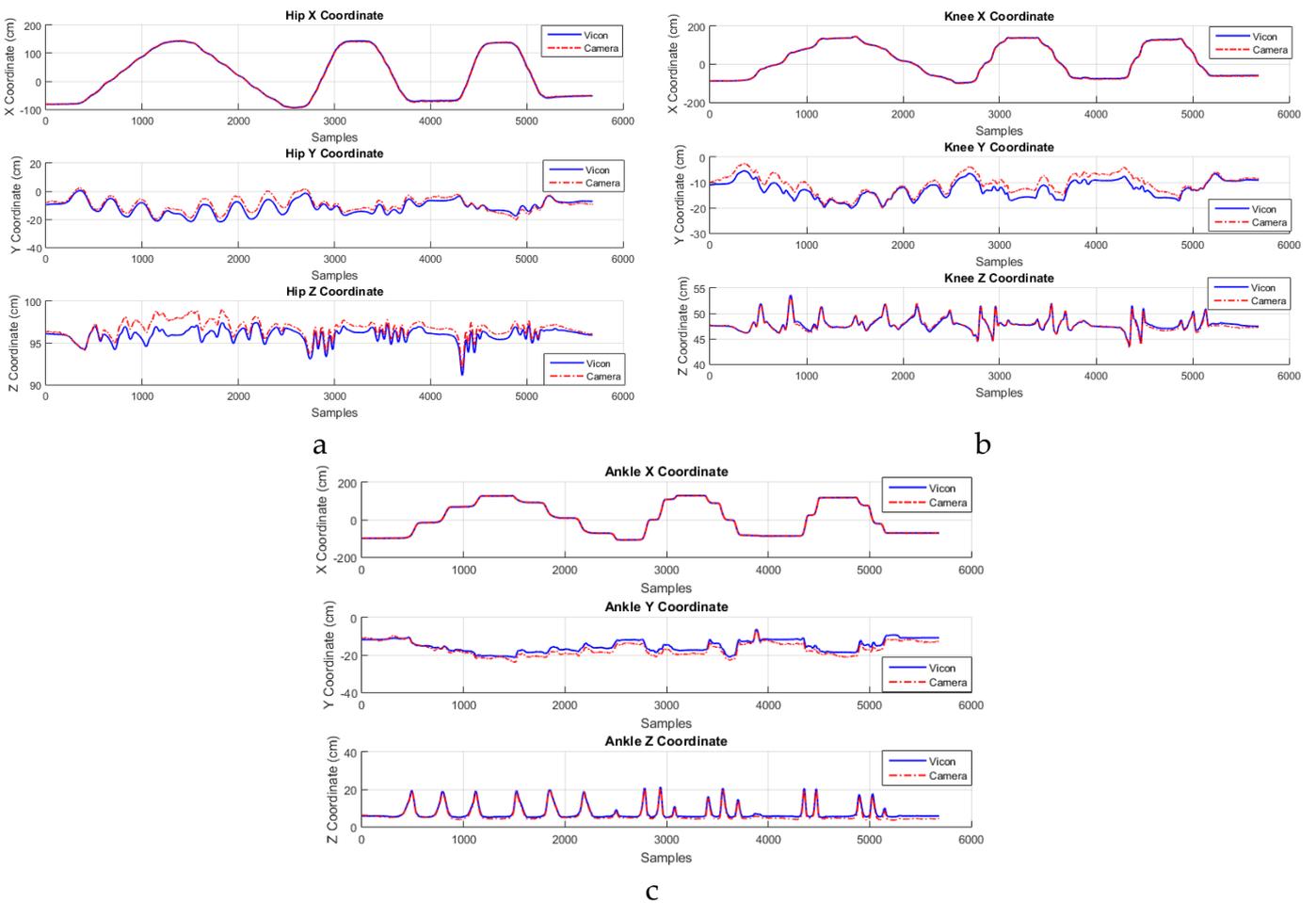

Figure 7. The results of a)Hip, b)Knee, c)Ankle joint motion tracking using camera-marker only – First Experiment

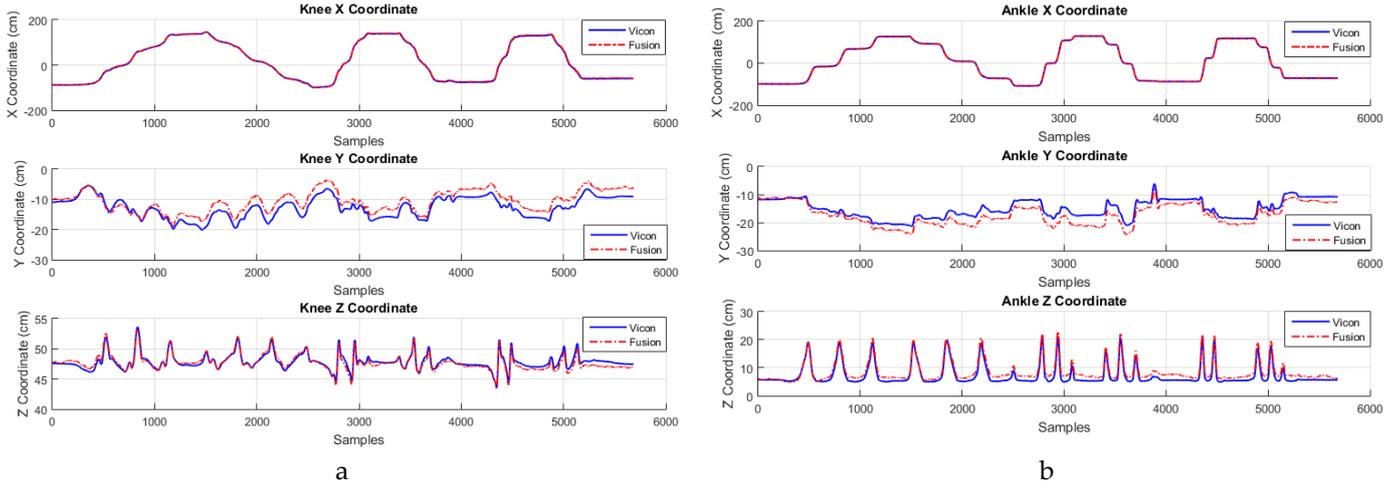

Figure 8. The results of a)Knee, b)Ankle joint motion tracking by fusion – First experiment

Table 1  The RMSE of leg joints' location tracked by IMUs only (Centimeter) - First Experiment

| Joint | $X$ | $Y$ | $Z$ | $\sqrt{X^2 + Y^2 + Z^2}$ |
|---|---|---|---|---|
| Ankle | 4.0 | 2.3 | 3.5 | 5.8 |
| Knee | 3.0 | 2.3 | 2.3 | 4.4 |
| Overall | 3.5 | 2.3 | 2.9 | 5.1 |

Table 2  The RMSE of leg joints' 2D location tracked by camera (Centimeter) – First Experiment

| Joint | $X$ | $Z$ | $\sqrt{X^2 + Z^2}$ |
|---|---|---|---|
| Ankle | 1.7 | 1.5 | 2.3 |
| Knee | 1.5 | 1.4 | 2.1 |
| Hip | 1.1 | 1.0 | 1.5 |
| Overall | 1.5 | 1.3 | 2.0 |

Table 3  The RMSE of leg joints' depth estimation by camera (Centimeter) - First Experiment

| Joint | Hip | Knee | Ankle | Overall |
|---|---|---|---|---|
| $Y$ | 4.6 | 5.4 | 6.1 | 5.4 |

Table 4  The RMSE of leg joints' 2D location tracking- Fusion (Centimeter) – First Experiment

| Joint | $X$ | $Z$ | $\sqrt{X^2 + Z^2}$ | Error change relative to IMU |
|---|---|---|---|---|
| Ankle | 2.9 | 2.1 | 3.6 | −30% |
| Knee | 2.1 | 1.7 | 2.7 | −31% |
| Overall | 2.5 | 1.9 | 3.1 | −30% |

Table 5  The RMSE of leg joints' depth estimation by fusion (Centimeter) - First Experiment

| Joint | Ankle | Knee | Overall | Error change relative to camera |
|---|---|---|---|---|
| $Y$ | 5.9 | 5.2 | 5.6 | −3% |

From Figure 6, it can be seen that joint location integrated directly from gyroscope signals with a correction from accelerometer with magnetometer signals, does not have the drift problem, which shows the effectiveness of gradient descent method and kinematic constraint. However, looking at Table 1, the overall error of tracking by IMUs is not acceptable for ankle joint. A hypothesis is that ankle's higher acceleration relative to other two joints reduces its location estimation accuracy. Another hypothesis is an accumulation of both IMUs' error in calculating ankle joint's location.

The results from Table 2 and Figure 7 for the walking test show that the marker-camera system has a high accuracy in tracking 2D joint location during low-acceleration movements. Also, the proposed depth estimation algorithm has an acceptable accuracy in computing joints depth illustrated in Table 3. Due to the lower acceleration of hip joint, its location has a lower RMSE relative to other joints. Updating in each frame, the location of joints in the camera-marker system is completely drift-free.

Unlike the IMU-based tracking system in which hip joint was derived from the Vicon system, in fusion algorithm, the hip joint location is calculated by the camera marker system. The results for the location of leg joints calculated by the proposed fusion method are similar to those of the Vicon system, as shown in Figure 8. It can be seen in Table 4 and Table 5 that in the fusion algorithm, the RMSE was reduced compared to both IMU and camera systems alone. All in all, the accuracy of proposed fusion algorithm is high in tracking joint location during low acceleration movements.

*6.2.2. Second Experiment*

In the second stage, the subject was asked to run in his place for about 20 seconds. Due to the high acceleration of IMUs in this test, shown in Figure 9, the magnetometer and accelerometer's data could not be used to reduce drift in gyroscopes.

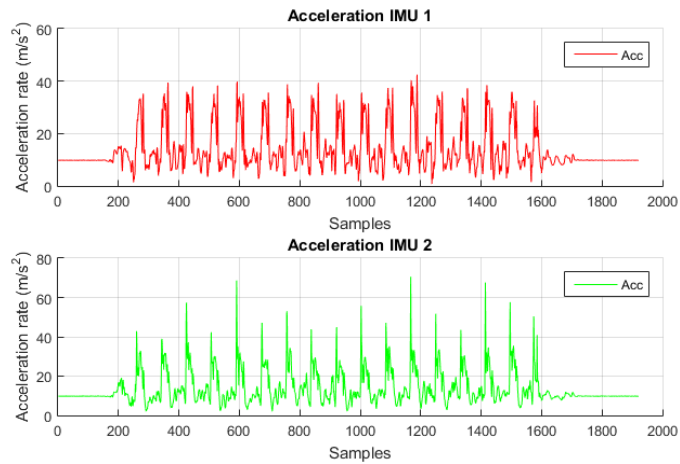

**Figure 9. Acceleration of IMUs in the second experiment**

Results of IMU sensors, camera-marker, and the fusion algorithm is shown respectively in Figure 10, Figure 11 and Figure 12 compared to the Vicon system. The accuracy results of each algorithm are summarized in Table 6 to Table 10. These tables show the mean RMSEs of joints location.

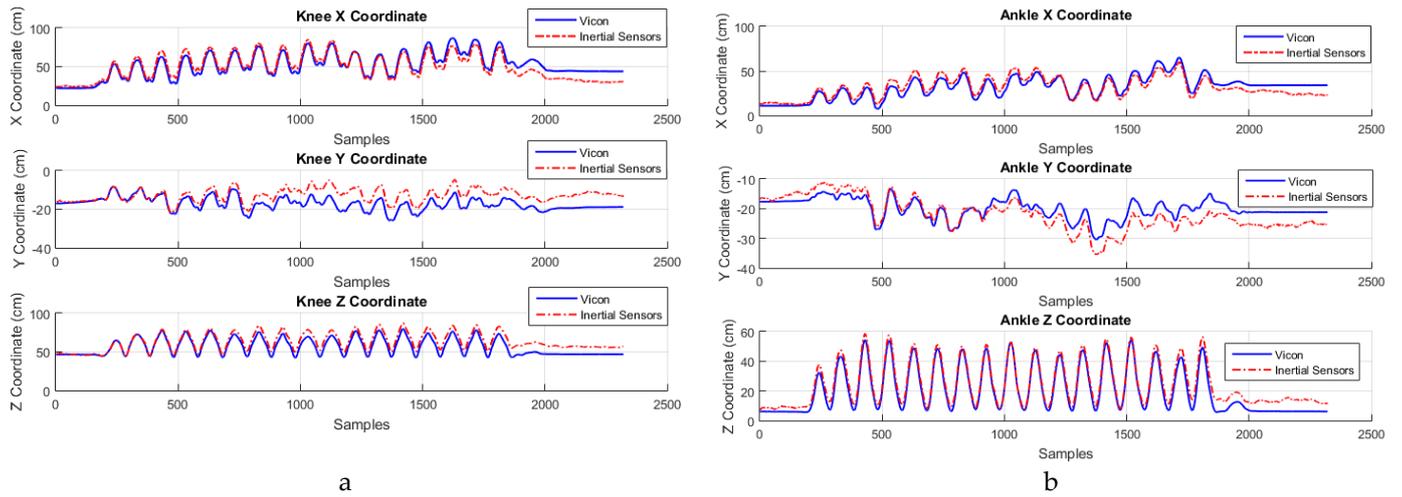

Figure 10. The results of a)Knee, b)Ankle joint motion tracking using IMUs only – Second experiment

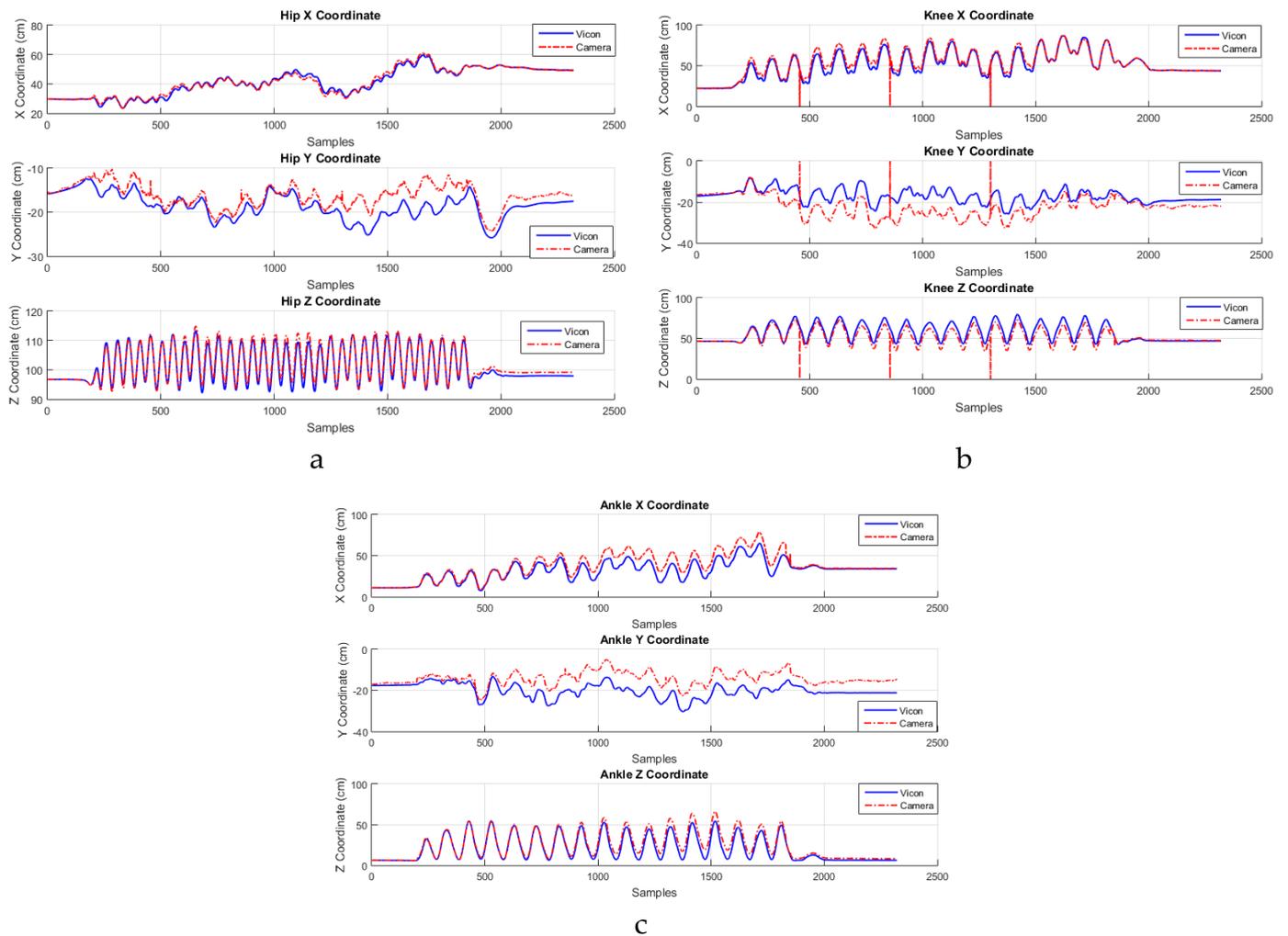

Figure 11. The results of a)Hip, b)Knee, c)Ankle joint motion tracking using camera-marker only – Second Experiment

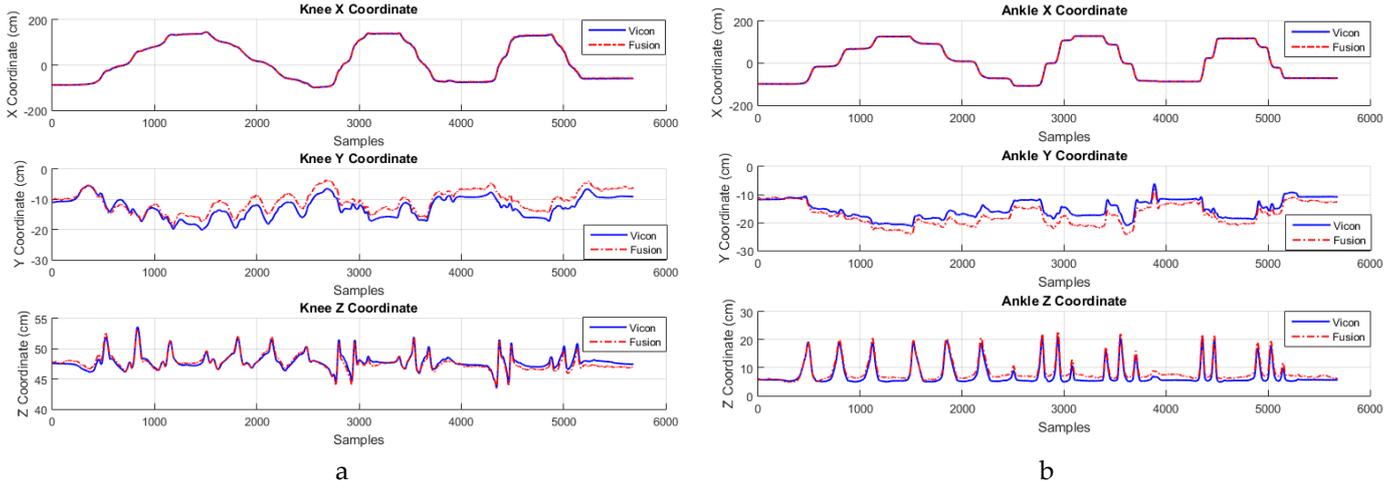

Figure 12. The results of a)Knee, b)Ankle joint motion tracking by fusion – Second experiment

Table 6  The RMSE of leg joints' location tracked by IMUs only (Centimeter) - Second Experiment

| Joint | $X$ | $Y$ | $Z$ | $\sqrt{X^2+Y^2+Z^2}$ |
|---|---|---|---|---|
| Ankle | 5.0 | 3.4 | 4.0 | 7.2 |
| Knee | 6.9 | 3.8 | 4.6 | 9.1 |
| Overall | 6.0 | 3.6 | 4.3 | 8.2 |

Table 7  The RMSE of leg joints' 2D location tracked by camera (Centimeter) – Second Experiment

| Joint | $X$ | $Z$ | $\sqrt{X^2+Z^2}$ |
|---|---|---|---|
| Ankle | 4.4 | 4.7 | 6.4 |
| Knee | 5.3 | 5.8 | 7.9 |
| Hip | 2.0 | 2.3 | 3.0 |
| Overall | 4.1 | 4.5 | 6.1 |

Table 8  The RMSE of leg joints' depth estimation by camera (Centimeter) - Second Experiment

| Joint | Hip | Knee | Ankle | Overall |
|---|---|---|---|---|
| $Y$ | 7.4 | 9.5 | 8.1 | 8.3 |

Table 9  The RMSE of leg joints' 2D location tracking- Fusion (Centimeter) – Second Experiment

| Joint | $X$ | $Z$ | $\sqrt{X^2+Z^2}$ | Error change relative to IMU |
|---|---|---|---|---|
| Ankle | 4.3 | 3.7 | 5.6 | −10% |
| Knee | 5.7 | 4.2 | 7.0 | −12% |
| Overall | 5.1 | 4.0 | 6.4 | −11% |

Table 10  The RMSE of leg joints' depth estimation by fusion (Centimeter) - Second Experiment

| Joint | Ankle | Knee | Overall | Error change relative to camera |
|---|---|---|---|---|
| $Y$ | 7.8 | 9.0 | 8.4 | −5% |

Due to the high-acceleration of movements, drift was presented in motion tracking by IMUs-only which can be seen in Figure 10. Also, overall RMSEs of tracking knee and ankle joints were increased compared to the first experiment (Table 6). Unlike the first experiment, in the second experiment, knee joint RMSE is higher than ankle joint. The possible reason is the higher magnitude of the acceleration of this joint compared to the ankle joint during the in-place running.

The RMSEs of Camera-marker system in Table 7 and Table 8, as well as IMU-based system, have increased in the high-acceleration movements compared to the low-acceleration one, However, as it is illustrated in Figure 11, this system dose not have drift problem. This is because the system updates its results after each frame. Comparing RMSE of 2D location of joints in Table 7, it can be seen that hip joint has a relatively lower RMSE in compare to other joints because of its lower acceleration magnitude. Thus, hip joint location is tracked with a high accuracy in both low and high acceleration movements. Depth estimation algorithm also has a lower accuracy in the second experiment. The possible sources of error in the second experiment are 1) low frequency of camera, and 2) displacement of markers on the leg skin. Due to missing knee joint's marker in some frames, there is a jump in its location curves shown in Figure 11.b. In these frames, the x and z locations of the joint are set to zero and its depth is computed using interpolation of other joints depth. If all joints' marker were missed, all joints' depth is set equal to zero.

Although the IMU-based system has drift problem, the fusion algorithm has reduced it significantly as it is shown in Figure 12 and Table 9. The effect of missed markers in some frames have been compensated by the fusion algorithm. The results show that depth estimation algorithm's RMSE has decreased after fusion. As it was pointed, hip joint location, which has a high accuracy in both high and low acceleration motions, is the base of calculating knee and ankle joints' location by IMUs which results in reducing joints' depth estimation by IMUs.

In general, in tracking motion by integrating IMUs and camera data, the errors were less than any of the sensors individually, which shows that the use of complementary properties of both systems has been effective in increasing motion tracking accuracy.

## *7. Conclusions*

This paper proposes a quaternion-based Extended Kalman filter approach to recover the human leg segments motions with a set of IMU sensors data fused with camera-marker system data. The suggested applications are for human motion monitoring and analysis in rehabilitation and sports medicine. Based on the complementary properties of the inertial sensors and the camera-marker system, in the introduced new measurement model, the orientation data of the upper leg and the lower leg is updated through three measurement equations. The positioning of the human body is made possible by the tracked position of the pelvis joint by the camera-marker system. A mathematical model had been utilized to estimate joints' depth in 2D images. The efficiency of the approach herein designed is demonstrated through two experiments evaluated by an optical motion tracker system. During the first stage of the test, the subject walked back and forth in the laboratory. The drift of gyroscopes had been reduced by magnetometer and accelerometer signals together with camera-marker system's data. Also, the depth estimation algorithm performed with high accuracy. During the second experiment, the subject ran while standing in his place in front of the camera. In this stage, updating in each frame, camera-marker data was reducing drift of the system. Moreover, the effect of missing markers in some frames was eliminated after the fusion algorithm. The obtained results and RMSE illustrate the performance of the proposed approach to estimate the human leg movements in both high and low acceleration motions with small errors. Future works will be focused on the application of the proposed approach in the online tracking of the human body. Also, the tracking of the whole human body based is intended.

*References*

1. Zhang, Z., Z. Huang, and J. Wu. *Hierarchical information fusion for human upper limb motion capture*. in *Information Fusion, 2009. FUSION'09. 12th International Conference on*. 2009. IEEE.

2. Sandau, M., et al., *Markerless motion capture can provide reliable 3D gait kinematics in the sagittal and frontal plane.* Medical engineering & physics, 2014. **36**(9): p. 1168-1175.

3. Sun, S., et al. *Adaptive Kalman filter for orientation estimation in micro-sensor motion capture*. in *Information Fusion (FUSION), 2011 Proceedings of the 14th International Conference on*. 2011. IEEE.

4. Sabatini, A.M., *Quaternion-based extended Kalman filter for determining orientation by inertial and magnetic sensing.* IEEE Transactions on Biomedical Engineering, 2006. **53**(7): p. 1346-1356.

5. Foxlin, E. *Inertial head-tracker sensor fusion by a complementary separate-bias Kalman filter*. in *Virtual Reality Annual International Symposium, 1996., Proceedings of the IEEE 1996*. 1996. IEEE.

6. Bachmann, E.R., *Inertial and magnetic tracking of limb segment orientation for inserting humans into synthetic environments*. 2000, Naval Postgraduate School Monterey CA.

7. Balan, A.O., L. Sigal, and M.J. Black. *A quantitative evaluation of video-based 3D person tracking*. in *2005 IEEE International Workshop on Visual Surveillance and Performance Evaluation of Tracking and Surveillance*. 2005. IEEE.

8. Madgwick, S., *An efficient orientation filter for inertial and inertial/magnetic sensor arrays.* Report x-io and University of Bristol (UK), 2010.

9. Grewal, M.S. and A.P. Andrews, *Kalman filtering: theory and practice*. 1993, Prentice Hall Englewood Cliffs, New Jersey.

10. Simon, D., *Optimal state estimation: Kalman, H infinity, and nonlinear approaches*. 2006: John Wiley & Sons.

11. Pourtakdoust, S. and H. Ghanbarpour Asl, *An adaptive unscented Kalman filter for quaternion-based orientation estimation in low-cost AHRS.* Aircraft Engineering and Aerospace Technology, 2007. **79**(5): p. 485-493.

12. Nützi, G., et al., *Fusion of IMU and vision for absolute scale estimation in monocular SLAM.* Journal of intelligent & robotic systems, 2011. **61**(1): p. 287-299.

13. Araguás, G., et al., *Quaternion-based orientation estimation fusing a camera and inertial sensors for a hovering UAV.* Journal of Intelligent & Robotic Systems, 2015. **77**(1): p. 37.

14. Zhou, H., H. Hu, and Y. Tao, *Inertial measurements of upper limb motion.* Medical and Biological Engineering and Computing, 2006. **44**(6): p. 479-487.

15. Cutti, A.G., et al., *Ambulatory measurement of shoulder and elbow kinematics through inertial and magnetic sensors.* Medical & biological engineering & computing, 2008. **46**(2): p. 169-178.

16. Yun, X. and E.R. Bachmann, *Design, implementation, and experimental results of a quaternion-based Kalman filter for human body motion tracking.* IEEE transactions on Robotics, 2006. **22**(6): p. 1216-1227.

17. Atrsaei, A., H. Salarieh, and A. Alasty, *Human Arm Motion Tracking by Orientation-Based Fusion of Inertial Sensors and Kinect Using Unscented Kalman Filter.* Journal of biomechanical engineering, 2016. **138**(9): p. 091005.

18. Madgwick, S.O., A.J. Harrison, and R. Vaidyanathan. *Estimation of IMU and MARG orientation using a gradient descent algorithm*. in *2011 IEEE International Conference on Rehabilitation Robotics*. 2011. IEEE.

19. Grana, C., D. Borghesani, and R. Cucchiara, *Optimized block-based connected components labeling with decision trees.* IEEE Transactions on Image Processing, 2010. **19**(6): p. 1596-1609.

20. Chang, W.-Y., C.-C. Chiu, and J.-H. Yang, *Block-based connected-component labeling algorithm using binary decision trees.* Sensors, 2015. **15**(9): p. 23763-23787.

21. He, L., et al., *The connected-component labeling problem: A review of state-of-the-art algorithms.* Pattern Recognition, 2017. **70**: p. 25-43.

22. Deepu, R., S. Murali, and V. Raju. *A Mathematical model for the determination of distance of an object in a 2D image*. in *Proceedings of the International Conference on Image Processing, Computer Vision, and Pattern Recognition (IPCV)*. 2013. The Steering Committee of The World Congress in Computer Science, Computer Engineering and Applied Computing (WorldComp).

23. Zhang, Z.-Q. and J.-K. Wu, *A novel hierarchical information fusion method for three-dimensional upper limb motion estimation.* IEEE transactions on instrumentation and measurement, 2011. **60**(11): p. 3709-3719.

24. Abyarjoo, F., et al., *Monitoring human wrist rotation in three degrees of freedom.* Southeastcon, 2013 Proceedings of IEEE, IEEE, 2013: p. 1-5.

25. Davenport, P.B., *Attitude determination and sensor alignment via weighted least squares affine transformations.* 1971.